\documentclass[journal]{IEEEtran}
\usepackage{subfigure}
\usepackage{graphicx}
\usepackage{amsmath,amsfonts}
\usepackage{cite}

\ifCLASSINFOpdf
\else
\fi
\begin{document}
%
\title{Task-Specific Data Preparation for Deep Learning to Reconstruct Structures of Interest from Severely Truncated CBCT Data}
%
%
%

\author{Yixing~Huang$^\ast$, Fuxin~Fan$^\ast$, Ahmed~Gomaa, Andreas~Maier, Rainer~Fietkau, Christoph~Bert, and Florian~Putz
\thanks{Y. Huang, A. Gomaa, R. Fietkau, C. Bert, and F. Putz are with Department of Radiation Oncology, University Hospital Erlangen, Friedrich-Alexander-Universit\"at Erlangen-N\"urnberg, 91054 Erlangen, Germany. They are also with Comprehensive Cancer Center Erlangen-EMN (CCC ER-EMN), 91054 Erlangen, Germany. }
\thanks{F. Fan and A. Maier are with Pattern Recognition Lab, Friedrich-Alexander-Universit\"at Erlangen-N\"urnberg, 91058 Erlangen, Germany.}
\thanks{Y. Huang and F. Fan contribute equally. Correspondence: yixing.yh.huang@fau.de, fuxin.fan@fau.de}
\thanks{Published in the proceeding of CT-Meeting 2024.}
}

\maketitle

\begin{abstract}
Cone-beam computed tomography (CBCT) is widely used in interventional surgeries and radiation oncology. Due to the limited size of flat-panel detectors, anatomical structures might be missing outside the limited field-of-view (FOV), which restricts the clinical applications of CBCT systems. Recently, deep learning methods have been proposed to extend the FOV for multi-slice CT systems. However, in mobile CBCT system with a smaller FOV size, projection data is severely truncated and it is challenging for a network to restore all missing structures outside the FOV. In some applications, only certain structures outside the FOV are of interest, e.g., ribs in needle path planning for liver/lung cancer diagnosis. Therefore, a task-specific data preparation method is proposed in this work, which automatically let the network focus on structures of interest instead of all the structures. Our preliminary experiment shows that Pix2pixGAN with a conventional training has the risk to reconstruct false positive and false negative rib structures from severely truncated CBCT data, whereas Pix2pixGAN with the proposed task-specific training can reconstruct all the ribs reliably. The proposed method is promising to empower CBCT with more clinical applications.
\end{abstract}


%
\IEEEpeerreviewmaketitle

\section{Introduction}
%
%
%
%
\IEEEPARstart{C}{one-beam} computed tomography (CBCT) imaging is widely used in interventional surgeries \cite{orth2009c,fan2022fiducial} and radiation oncology \cite{bapst2016cone,fonseca2021evaluation,karius2022first}. Data truncation is a common problem in practical CBCT applications, which arises in two scenarios: a) Application of collimators in interior tomography \cite{yu2009compressed,han2019one,huang2021data} to reduce dose exposure to sensitive organs, e.g., protecting eyes in head imaging; b) The flat-panel detectors with a limited size are not large enough to cover entire patients or additional structures of interest (SOI) \cite{fonseca2021evaluation,bapst2016cone,fan2022fiducial}, e.g., distorted fiducial markers in spine surgery and missing shoulders in head $\&$ neck adaptive radiotherapy. To extend the field of view (FOV) in the latter scenario, special scan trajectories have been developed, including traverse-continuous-rotate scan \cite{nahamoo1981design}, elliptical trajectory \cite{li2010novel}, and multiple source-translation scan \cite{yu2023analytical}. Nevertheless, the circular trajectory is the most widely applied trajectory in practice. 

To extend the FOV of CBCT with a conventional circular trajectory, conventional extrapolation methods such as cosine function fitting \cite{pmid15702338} and water cylinder extrapolation (WCE) \cite{Hsieh2004WCE} have been applied. Compressed sensing techniques are effective to reduce cupping artifacts inside the FOV, but are not sufficient to restore anatomical structures outside the FOV \cite{yu2009compressed,huang2021data}. Recently, the advent of advanced deep learning techniques has opened up new possibilities for extending the FOV \cite{fonseca2021evaluation,huang2021data,khural2022deep,xu2023body,xie2023inpainting}, which has the potential to empower CBCT with a broader range of clinical applications.  Fonseca \textit{et al.} has demonstrated the superiority of a U-Net based deep learning method called HDeepFov to a recently released Siemens commercial algorithm HDFov \cite{fonseca2021evaluation}. In our previous work \cite{huang2021data}, a general plug-and-play framework has been proposed to extend the FOV from projection data extrapolated from deep learning prior reconstruction images, which guarantees data consistency inside the FOV and makes the most of the deep learning models to estimate structures outside the FOV. Xu \textit{et al.} have proposed a two-step method for body composition analysis \cite{xu2023body}: one network for FOV border extension and the other network for missing tissue imputation. Xie \textit{et al.} has demonstrated that CT images inpainted by their proposed gated convolutional (GatedConv) network can achieve more accurate dose planning than other networks for radiotherapy \cite{xie2023inpainting}.

Note that multi-slice CT systems are considered in the previous work \cite{fonseca2021evaluation,khural2022deep,xu2023body,xie2023inpainting}, which typically have a larger FOV size than CBCT systems. Especially, mobile C-arm CBCT systems can have a very small FOV of 16\,cm in diameter \cite{fan2022fiducial}. As a consequence, a large portion of anatomical structures are severely truncated, posing a challenge for neural networks to estimate all missing structures. In some clinical applications, only certain structures outside the FOV are of interest, for example, fiducial markers in spine surgery \cite{fan2022fiducial} and skeletal structures for path planning in image-guided needle biopsy in lung/liver cancer diagnosis \cite{de2016image,bapst2016cone}. In this work, a task-specific data preparation method for deep learning to reconstruct SOI outside FOV from severely truncated CBCT data is proposed.

\section{Methods}

\subsection{Task-Specific Data Preparation}
In the scenarios with severely truncated data, it is very challenging for a deep learning model to restore complete structures outside the FOV accurately due to the large amount of missing data. In this work, we propose a task-specific learning strategy to let neural networks focus on reconstructed SOI only, since only certain structures outside the FOV are of interest in certain applications.

\subsubsection{Conventional data preparation for FOV extension}
 To prepare training data, it is straightforward to use images reconstructed from truncated data as the neural network input and images reconstructed from untruncated (complete) data as the neural network output \cite{huang2020field,xu2023body,xie2023inpainting}. Such a conventional data preparation way can be represented as follows,
 \begin{equation}
\begin{array}{c}
   \boldsymbol{f}_\text{input} = \mathcal{R} (\boldsymbol{A}_{\text{TP}}\cdot\boldsymbol{f}), \\
       \boldsymbol{f}_\text{label, conventional} = \mathcal{R} (\boldsymbol{A}_{\text{UTP}}\cdot \boldsymbol{f}),
\end{array}
\label{eqn:conventionalDataPrepation}
\end{equation}
 where $\boldsymbol{f}$ is the complete reference 3D image, $\boldsymbol{f}_\text{input}$ is the network input, and $\boldsymbol{f}_\text{label, conventional}$ is the conventional network output. The operation $\mathcal{R}$ means image reconstruction, which is FDK reconstruction from water cylinder extrapolated data (WCE) in particular in this work \cite{huang2021data}. $\boldsymbol{A}_{\text{TP}}$ and $\boldsymbol{A}_{\text{UTP}}$ are paired truncated and untruncated forward projection operators. In practice, it is challenging to acquire paired truncated and untruncated projection data in real CBCT systems. As deep learning models trained by realistic synthetic data can generalize to real data \cite{gao2023synthetic}, paired truncated and untruncated forward projection operators (and data) are easily accessible via simulation.

\subsubsection{Task-specific data preparation for FOV extension}
To enable neural networks to concentrate on the SOI, a possible method is to assign greater importance to these structures compared to others through increased weighting. However, in images that are reconstructed from incomplete data, artifacts such as streaks exist globally. As a result, segmenting these areas for weight application becomes impractical. Moreover, if the weighting is restricted solely to the SOI region, the artifacts originating from the SOI will persist outside this specifically weighted zone.

 To avoid such difficulty, a task-specific data preparation has been successfully applied to fiducial marker recovery in navigation assisted spine surgery with severely truncated data \cite{fan2022fiducial}. In this work, the task-specific data preparation strategy is extended to the following general form:
\begin{equation}
\begin{array}{c}
   \boldsymbol{f}_\text{input} = \mathcal{R} (\boldsymbol{A}_{\text{TP}}\cdot\boldsymbol{f}) = \mathcal{R} (\boldsymbol{A}_{\text{TP}}(\boldsymbol{f}_{\text{Others}} ) + \mathcal{R} (\boldsymbol{A}_{\text{TP}}(\boldsymbol{f}_{\text{SOI}})),  \\
      \boldsymbol{f}_\text{label, task-specific} = \mathcal{R} (\boldsymbol{A}_{\text{TP}}(\boldsymbol{f}_{\text{Others}})) + \mathcal{R} (\boldsymbol{A}_{\text{UTP}}(\boldsymbol{f}_{\text{SOI}})),
\end{array}
\label{eqn:taskSpecificDataPreparation}
\end{equation}
where $\boldsymbol{f}_{\text{SOI}}$ denotes the segmented SOI and $\boldsymbol{f}_{\text{others}}$ denotes the other structures. Therefore, our task-specific data preparation relies on the segmentation of SOI as a prerequisite.

Compared with Eqn.\,(\ref{eqn:conventionalDataPrepation}), the inputs of both data preparations are fundamentally the same. The complete reconstruction can be partitioned into two parts when a linear reconstruction operator like FDK is used. In Eqn.\,(\ref{eqn:taskSpecificDataPreparation}), only the SOI are reconstructed from untruncated data in the neural network output. With such data preparation, the difference between an input image and its corresponding output image is originated from SOI only. Therefore, neural networks only need to focus on learning such difference. The task-specific learning makes the network neglect unimportant structures and focus on SOI only, hence it has the potential to generate more robust predictions for complex clinical data than conventional data preparation.

\begin{figure}[htb!]
    \centering
      \begin{minipage}{0.47\linewidth}
    \subfigure[Reference]{
    \includegraphics[width=\linewidth]{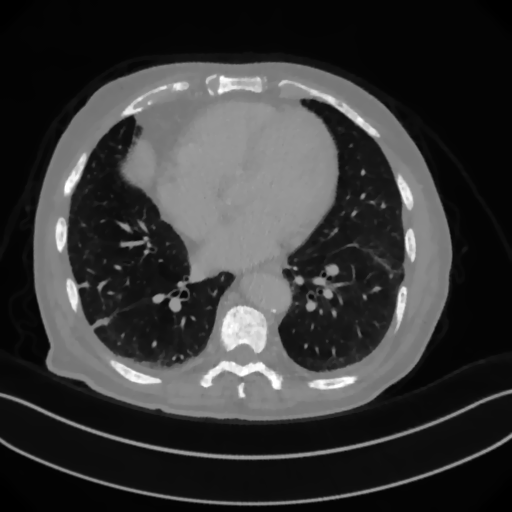}
    \label{subfig:Reference}
    }
    \end{minipage}
    \begin{minipage}{0.47\linewidth}
    \subfigure[Input/truncated reconstruction]{
    \includegraphics[width=\linewidth]{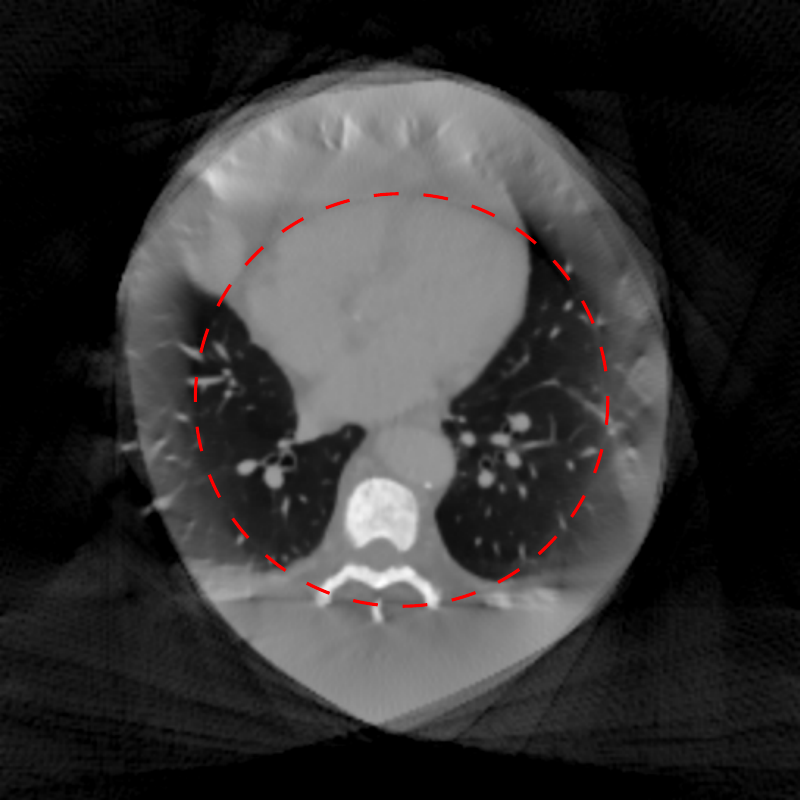}
    \label{subfig:ribInput1}
    }
    \end{minipage}
    
        \begin{minipage}{0.47\linewidth}
    \subfigure[Segmented ribs]{
    \includegraphics[width=\linewidth]{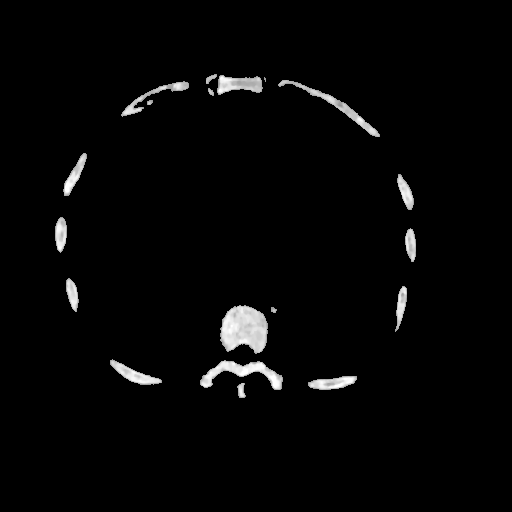}
    \label{subfig:ribSegment}
    }
    \end{minipage}
    \begin{minipage}{0.47\linewidth}
    \subfigure[Other anatomical structures]{
    \includegraphics[width=\linewidth]{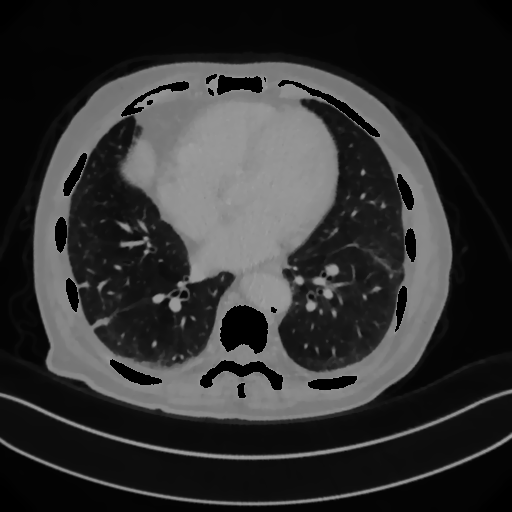}
    \label{subfig:otherSegment}
    }
    \end{minipage}

    \begin{minipage}{0.47\linewidth}
    \subfigure[Truncated reconstruction of ribs]{
    \includegraphics[width=\linewidth]{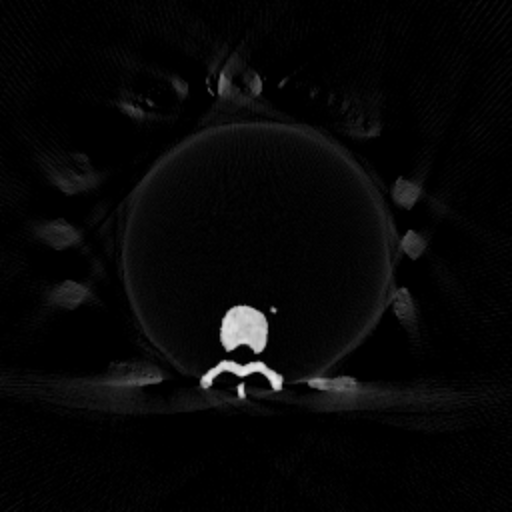}
    \label{subfig:ribReconstructionTruncat}
    }
    \end{minipage}
        \begin{minipage}{0.47\linewidth}
    \subfigure[Truncated reconstruction of others]{
    \includegraphics[width=\linewidth]{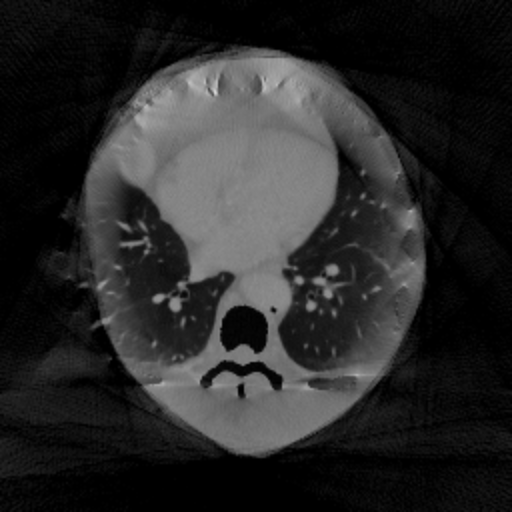}
    \label{subfig:othersReconstructionTruncat}
    }
    \end{minipage}

    \begin{minipage}{0.47\linewidth}
    \subfigure[Conventional label for training]{
    \includegraphics[width=\linewidth]{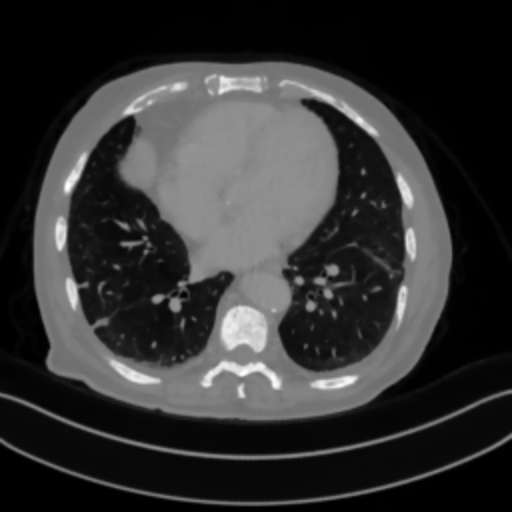}
    \label{subfig:ribReference}
    }
    \end{minipage}
   \begin{minipage}{0.47\linewidth}
    \subfigure[Task-specific label for training]{
    \includegraphics[width=\linewidth]{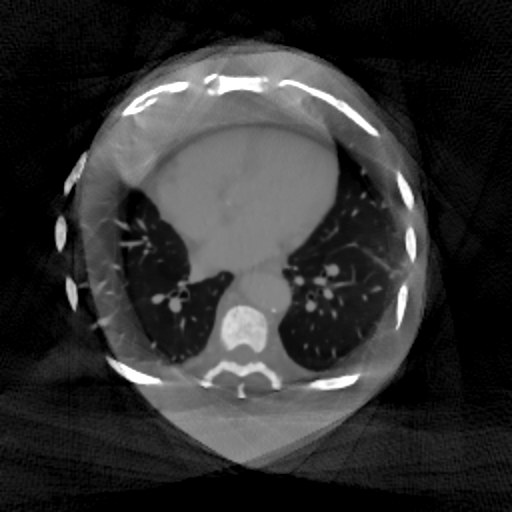}
    \label{subfig:ribTSoutput}
    }
    \end{minipage} 
     
 \caption{Intermediate reconstructions under different data preparations. (a) is the reference slice with all structures. (b) is the input reconstruction from truncated data. (c) and (d) contain ribs and other anatomical structures, respectively. (e) and (f) are truncated reconstruction for ribs and other anatomical structures, respectively. (g) shows the label by conventional data preparation. (h) is the label by task-specific data preparation.}   
  \label{Fig:RibDataPrep}
\end{figure}

\subsubsection{Exemplification with rib reconstruction in image-guided needle biopsy}
To show the generalizability of the task-specific learning for image reconstruction tasks from severely truncated data, potential applications to image-guided needle biopsy in lung cancer diagnosis \cite{de2016image} and interventional oncology of liver \cite{bapst2016cone} are considered, in addition to navigation assisted spine surgery \cite{fan2022fiducial}. In such applications, a 3D volume is needed to establish the safest needle path to the target lesion, where impenetrable obstacles like ribs need to be avoided. Nowadays such a 3D volume is typically acquired by multi-slice CT systems instead of CBCT systems because a large FOV is necessary to cover the lung/liver and ribs. CBCT systems with a small FOV can visualize the target region with high image quality, but the ribs are located outside the FOV, as displayed in the example of Fig.\,\ref{subfig:ribInput1}. 

\begin{figure}[htb!]
    \centering
         \begin{minipage}{\linewidth}
    \subfigure[Projection of ribs]{
    \includegraphics[width=\linewidth]{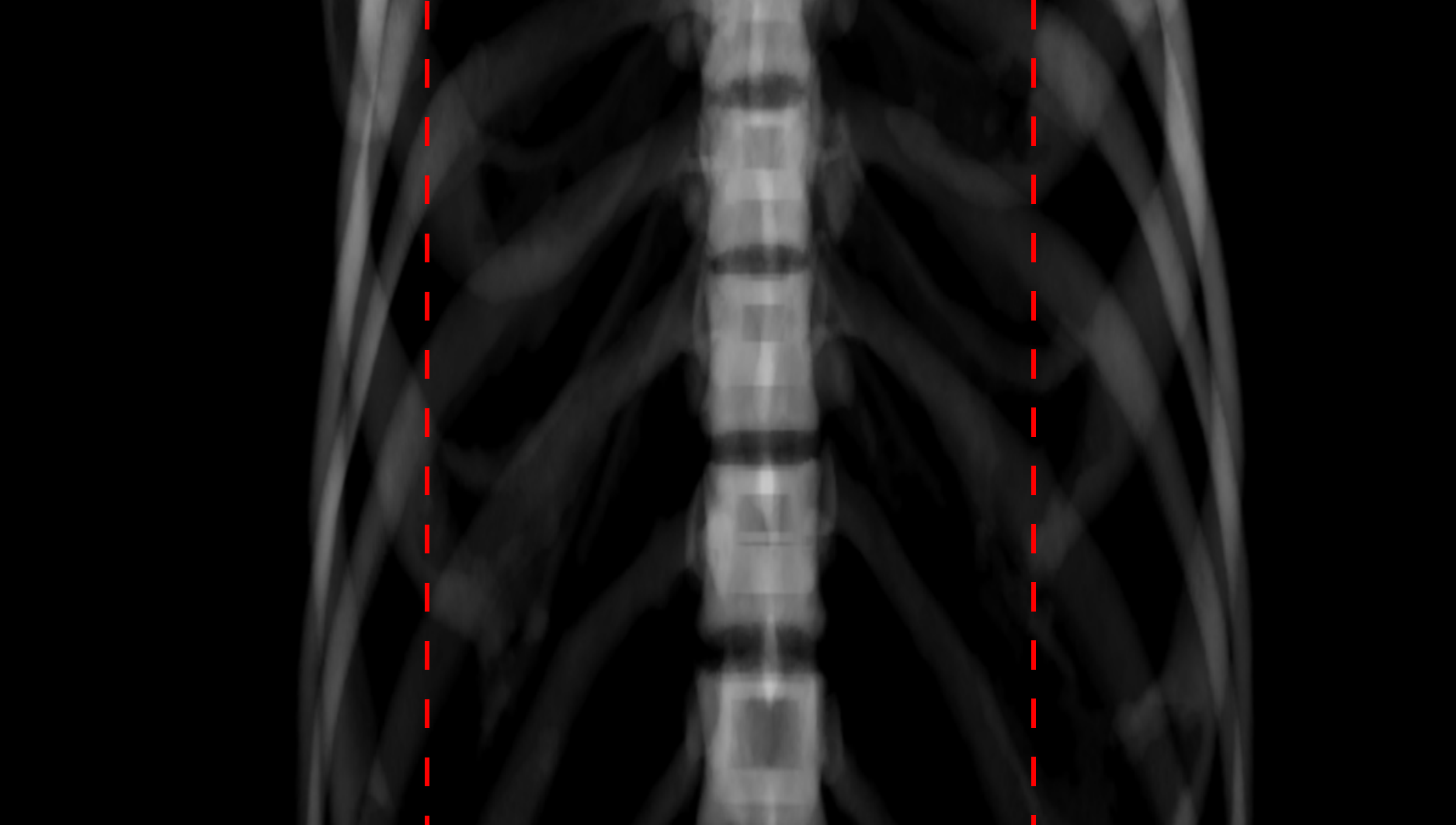}
    \label{subfig:ribProjection}
    }
    \end{minipage}
    \begin{minipage}{\linewidth}
    \subfigure[Projection of other anatomical structures]{
    \includegraphics[width=\linewidth]{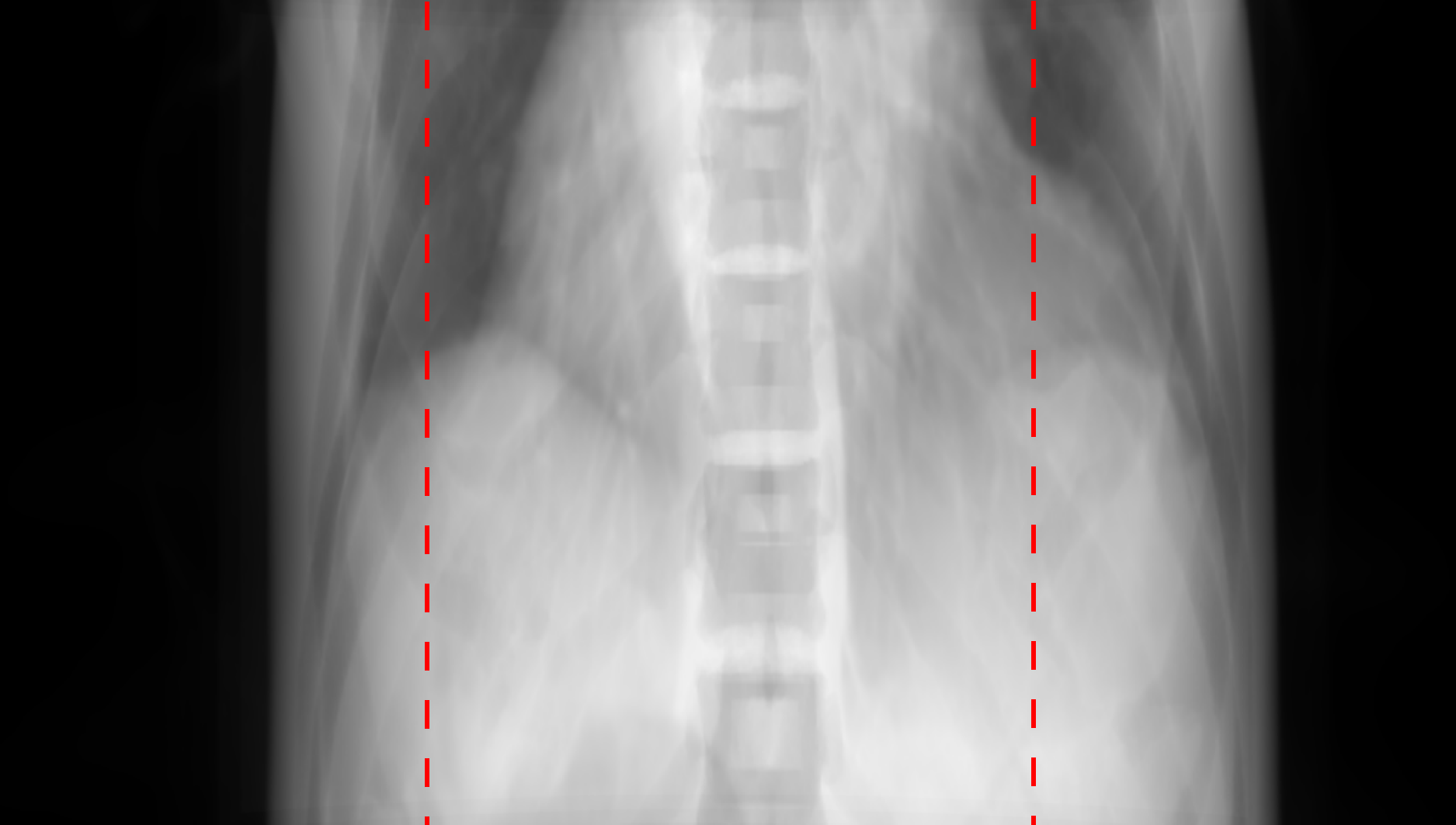}
    \label{subfig:othersProjection}
    }
    \end{minipage}
 
 \caption{The projections of one patient's ribs solely (a) and other anatomical structures (b), where the dashed lines indicate the truncation boundaries, display windows [0, 4.5] for (a) and [0, 9] for (b).}   
  \label{Fig:projections}
\end{figure}

With our task-specific learning, CBCT systems with a small FOV are potentially capable of such an application. To achieve this goal, the skeletal structures including ribs, vertebra and scapula of CT volumes are segmented first for training, as displayed in Fig.\,\ref{subfig:ribSegment}. The soft tissues are displayed in Fig.\,\ref{subfig:otherSegment}. Their projections are displayed in Fig.\,\ref{subfig:ribProjection} and Fig.\,\ref{subfig:othersProjection}, respectively, and their corresponding truncated reconstructions are displayed in Fig.\,\ref{subfig:ribReconstructionTruncat} (i.e., $\mathcal{R} (\boldsymbol{A}_{\text{TP}}(\boldsymbol{f}_{\text{SOI}}))$ in Eqn.\,(\ref{eqn:taskSpecificDataPreparation})) and Fig.\,\ref{subfig:othersReconstructionTruncat} (i.e., $\mathcal{R} (\boldsymbol{A}_{\text{TP}}(\boldsymbol{f}_{\text{Others}} )$ in Eqn.\,(\ref{eqn:taskSpecificDataPreparation})), respectively. In Fig.\,\ref{subfig:ribReconstructionTruncat}, the ribs outside the FOV are severely distorted with low contrast, but we can still see their presence. The conventional output for training is Fig.\,\ref{subfig:ribReference}, which is the reconstruction from nontruncated projection data. In contrast, the task-specific label for training is Fig.\,\ref{subfig:ribTSoutput}, which is the combination of Fig.\,\ref{subfig:ribSegment} and Fig.\,\ref{subfig:othersReconstructionTruncat}, whereas the input image Fig.\,\ref{subfig:ribInput} is the combination of Fig.\,\ref{subfig:ribSegment} and Fig.\,\ref{subfig:ribReconstructionTruncat}.


\subsection{Experimental Setup}
 To demonstrate this, the skeletal structures including ribs, vertebra and scapula of 18 AAPM Low Dose CT Grand Challenge volumes are segmented with 3D Slicer \cite{fedorov20123d} segmentation tools. A CBCT system with parameters in Tab.\,\ref{parameter} is used to simulate projection data. The simulated CBCT system parameters are aligned with a Siemens mobile C-arm system. Artificial Poisson noise is added considering a photo number of $10^6$ for each incident X-ray. 1360 slices from 17 volumes are used for training Pix2pixGAN models with either conventional data preparation or task-specific data preparation, and the 18th volume is used for test. The models are trained for 300 epochs with an initial learning rate of 0.001 and a decay rate of 0.97. The Adam optimizer is used. 
 
\begin{table}[htbp]
\begin{center}
\caption{{Parameters for the C-arm CBCT system and reconstruction.}
\label{parameter}
}
    \begin{tabular}{l|l}
    \hline
        \hline
        Parameter&Value\\
        \hline
         Scan angular range&200$^{\circ}$\\
        \hline
        Angular step&0.5$^{\circ}$\\
        \hline
        Source-to-detector distance& 1164 mm\\
        \hline
        Source-to-isocentor distance& 622 mm\\
        \hline
        Detector size& 500\,$\times$\,680\\
        \hline
        Extended virtual detector size& {1200}\,$\times$\,680\\
        \hline
        Detector pixel size (mm)& 0.608\,$\times$\,0.608\\
        \hline
         FOV diameter & {16.2\,cm}\\
       \hline 
        Reconstruction volume size&512\,$\times$\,512\,$\times$\,512\\
        \hline
        Reconstruction voxel size (mm)&0.625\,$\times$\,0.625\,$\times$\,0.625\\

        \hline
    \end{tabular}
\end{center}
\end{table}

\section{Results And Discussion}

\begin{figure}[htb!]
    \centering
    \begin{minipage}{0.47\linewidth}
    \subfigure[Reference]{
    \includegraphics[width=\linewidth]{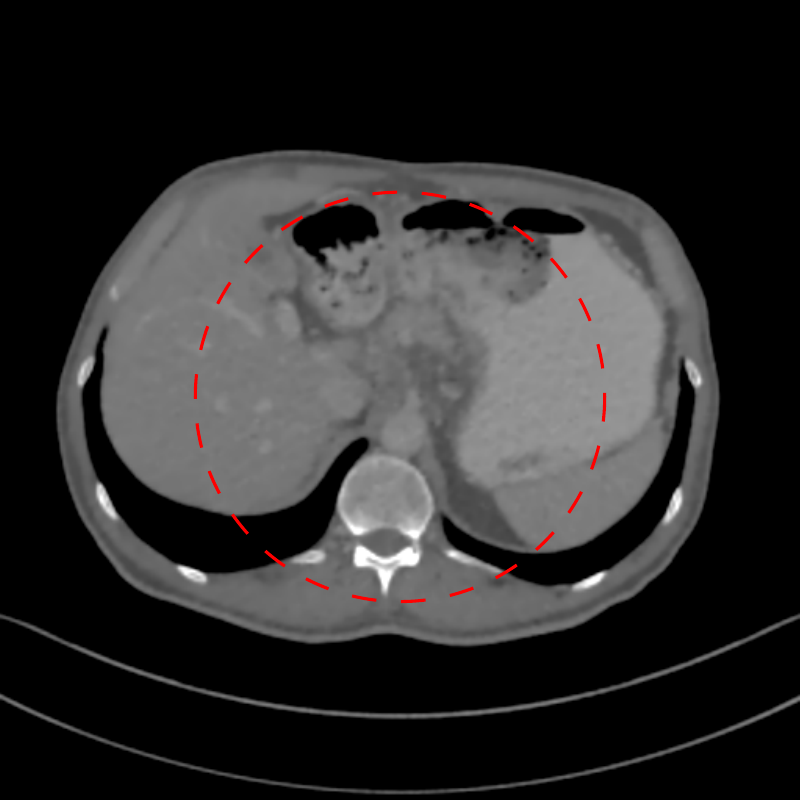}
    \label{subfig:ribGT}
    }
    \end{minipage}
    \begin{minipage}{0.47\linewidth}
    \subfigure[Input]{
    \includegraphics[width=\linewidth]{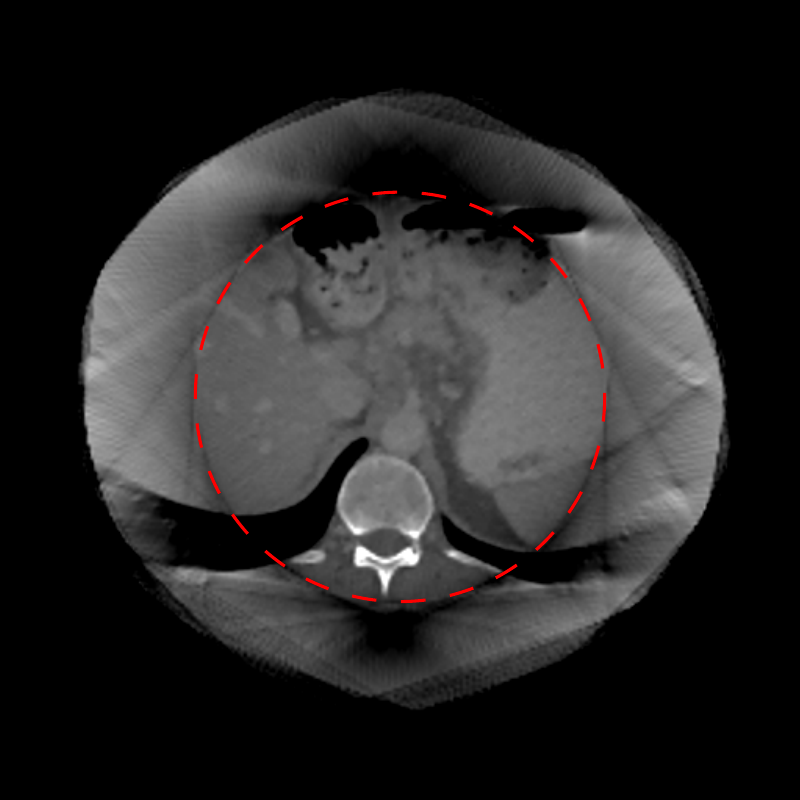}
    \label{subfig:ribInput}
    }
    \end{minipage}
   
   \begin{minipage}{0.47\linewidth}
    \subfigure[Pix2pixGAN, conventional]{
    \includegraphics[width=\linewidth]{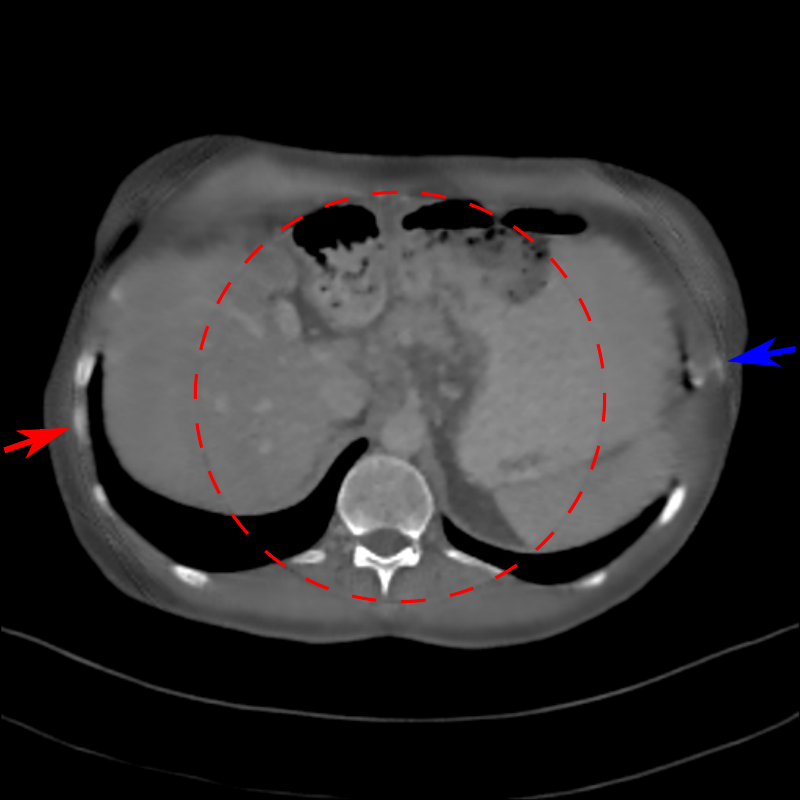}
    \label{subfig:ribReg}
    }
    \end{minipage} 
     \begin{minipage}{0.47\linewidth}
    \subfigure[Pix2pixGAN, task-specific]{
    \includegraphics[width=\linewidth]{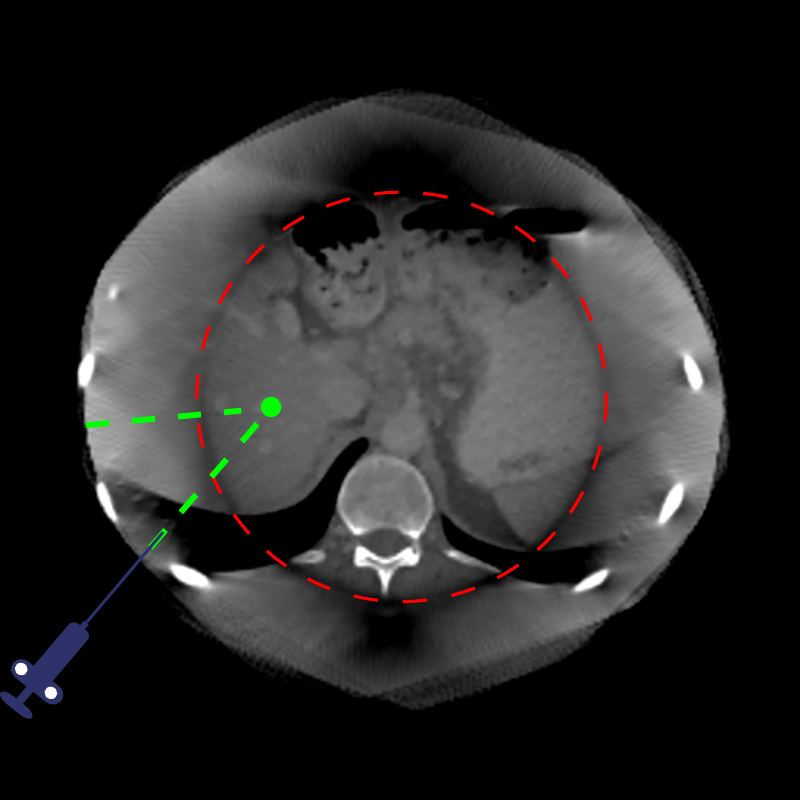}
    \label{subfig:ribTaskSp}
    }
    \end{minipage}
    
       \begin{minipage}{0.47\linewidth}
    \subfigure[Tasks-specific label]{
    \includegraphics[width=\linewidth]{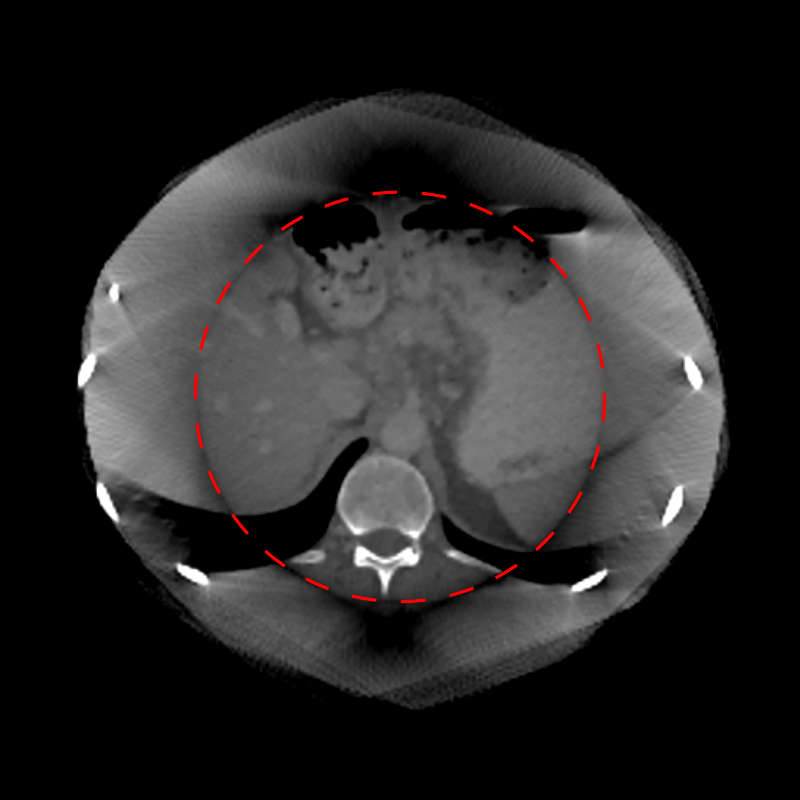}
    \label{subfig:taskSpecificLabel}
    }
    \end{minipage} 
     \begin{minipage}{0.47\linewidth}
    \subfigure[Difference, task-specific]{
    \includegraphics[width=\linewidth]{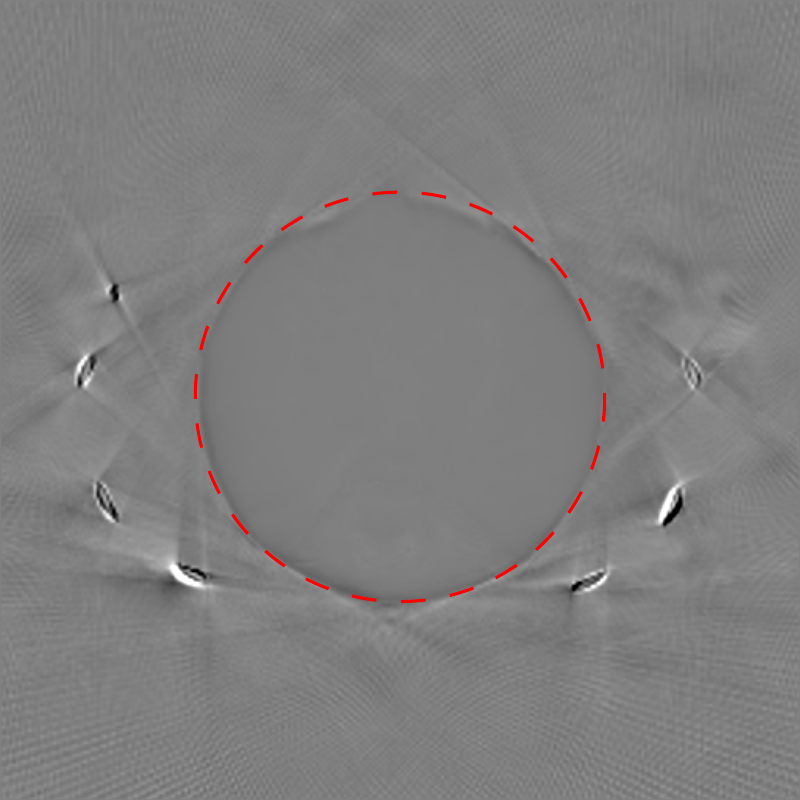}
    \label{subfig:ribTaskSpDiff}
    }
    \end{minipage}

 \caption{A potential application to rib reconstruction from severely truncated data for image-guided needle biopsy, where the structures inside the FOV (indicated by the dashed circle) and the ribs outside the FOV are of interest, window: [-600, 600]\,HU. (b) is the input image. (c) is the Pix2pixGAN output with conventional data preparation, where a false positive rib indicated by the red arrow is reconstructed and a rib indicated by the blue arrow has a low contrast and incorrect shape. (d) is the Pix2pixGAN output with our proposed task-specific data preparation, where the number of ribs is correct and the positions of ribs are accurate. The two green dash lines are two potential needle paths for the target (marked by the green point) between ribs. Its error image with respective the task-specific label (e) is displayed in (f).}   
  \label{Fig:RibRecons}
\end{figure}

To show the efficacy of the task-specific learning for image reconstruction from severely truncated data, the results of an exemplary slice are displayed in Fig.\,\ref{Fig:RibRecons}, where the structures inside the FOV and the ribs outside the FOV are of interest. In the input image (Fig.\,\ref{subfig:ribInput}), the ribs are almost not visible at all due to severe truncation. With the conventional data preparation, Pix2pixGAN reconstructs all the structures with a realistic appearance outside the FOV in Fig.\,\ref{subfig:ribReg}. As a result, the complete liver is visible. However, the body outline is inaccurate at the top left side. More importantly, the ribs of interest are reconstructed incorrectly. A false positive rib is reconstructed, as indicated by the red arrow in Fig.\,\ref{subfig:ribReg}. In addition, the rib indicated by the blue arrow has very low contrast and wrong shape. A binary rib mask is manually segmented from Fig.\,\ref{subfig:ribReg} and its Dice score is 0.906 with respect to the refence rib mask. In contrast, the ribs in the Pix2pixGAN output (Fig.\,\ref{subfig:ribTaskSp}) with our proposed task-specific learning are well reconstructed with a high Dice score of 0.958, where the number of ribs is correct and the positions are accurate, although their intensities are higher than those in the reference (Fig.\,\ref{subfig:ribGT}) (because the rib regions are overlapped with artifacts caused by the truncated reconstruction of other structures). With the known rib positions, two potential needle paths for the target (marked by the green point) can be found in Fig.\,\ref{subfig:ribTaskSp}.


\section{Conclusion}
In this work, a task-specific data preparation method is proposed to reconstruct SOI outside the FOV from severely truncated CBCT data. In the application to path planning in image-guided needle biopsy, Pix2pixGAN with the proposed method can reconstruct the number of ribs correctly and their locations accurately, whereas Pix2pixGAN with conventional training has the risk to reconstruct false positive and false negative ribs, although the images looks realistic. The proposed method has the potential for SOI reconstruction in other CT reconstruction tasks from severely insufficient data, e.g., extremely limited-angle and sparse-view tomography, which will be our future work.


%

\ifCLASSOPTIONcaptionsoff
  \newpage
\fi

\bibliographystyle{IEEEtran}
\bibliography{refs}
\end{document}